\documentclass[letterpaper, 10 pt, conference]{ieeeconf}  

\usepackage[
  expansion=alltext,
  protrusion=alltext-nott, 
  final 
]{microtype}
\usepackage{amsmath}
\usepackage{amssymb}
\usepackage{siunitx}
\usepackage{graphicx}
\usepackage{multirow}
\usepackage{booktabs}
\usepackage{placeins} 
\usepackage[nohyperlinks, nolist]{acronym}
\usepackage{bm}
\usepackage{color}
\usepackage[ruled,vlined]{algorithm2e}
\usepackage{makecell}
\usepackage[hidelinks]{hyperref}

\graphicspath{{figures}}
\definecolor{cmtgray}{gray}{0.6}

\newcommand*{\matr}[1]{\bm{#1}}
\newcommand*{\vect}[1]{\bm{#1}}

\newcommand*{\Tran}{^{\mathsf{T}}}

\newcommand*{\R}{\mathbb{R}}
\newcommand*{\Z}{\mathbb{Z}}
\newcommand*{\Sph}{\mathbb{S}}
\newcommand*{\SO}{\mathit{SO}}
\newcommand*{\SE}{\mathit{SE}}
\newcommand*{\se}{\mathfrak{se}}
\newcommand*{\so}{\mathfrak{so}}
\newcommand*{\increment}{\Delta}
\newcommand*{\tauoutc}{\tau_{\text{out,1}}} 
\newcommand*{\tauoutr}{\tau_{\text{out,2}}}
\newcommand*{\tauinc}{\tau_{\text{in,1}}}
\newcommand*{\tauinr}{\tau_{\text{in,2}}}

\acrodef{SDF}{signed distance field}
\acrodef{FFT}{fast Fourier transform}
\acrodef{CDF}{cumulative distribution function}
\acrodef{BVH}{bounding volume hierarchy}
\acrodef{AABB}{axis-aligned bounding box}
\acrodef{CMM}{coordinate measuring machine}
\acrodef{ICP}{iterative closest point}
\acrodef{CAD}{computer-aided design}

\IEEEoverridecommandlockouts

\title{\LARGE \bf
  From Swept Contact to Pose: Probe-Aware Registration via Complementary-Shape Docking
}

\author{Chen Chen, Yunwen Li, Yifan Xu, Xiangjie Yan, Chang Shu, Jianxia Hou, Shiji Song, and Xiang Li
  \thanks{C. Chen, Y. Xu, X. Yan, S. Song, and X. Li are with the Department of Automation, Tsinghua University.
    Y. Li is with Tsingscribe Medical Ltd., and also with D-MAVT, ETH Zürich.
    C. Shu and J. Hou are with Peking University School and Hospital of Stomatology.
    This work was supported
    in part by the Brain Science and Brain-like Intelligence Technology-National Science and Technology Major Project under Grant 2021ZD0201404,
    in part by the National Natural Science Foundation of China under Grant 62461160307,
    in part by the Fundamental and Interdisciplinary Disciplines Breakthrough Plan of the Ministry of Education of China under Grant JYB2025XDXM208,
    and in part by the Institute for Guo Qiang, Tsinghua University.
    Corresponding author: Xiang Li (xiangli@tsinghua.edu.cn)}%
}

\begin{document}

\maketitle
\thispagestyle{plain}
\pagestyle{plain}


\begin{abstract}
  Accurate registration between a prior model and the real scene is essential for high-precision robotic manipulation, yet optical methods suffer from long calibration chains, line-of-sight constraints, and fabrication errors.
  We propose a calibration-free alternative that reformulates contact registration as complementary-shape docking between the object and the probe's swept volume, explicitly accounting for probe geometry and leveraging both contact and non-contact evidence.
  Our solver integrates a global-to-local search via 3D FFT correlation over low-discrepancy \textit{SO}(3) samples, then followed by continuous \textit{SE}(3) refinement using Lie-algebra updates and analytic contact sensitivities.
  This pipeline yields efficient exploration and metric-grade convergence without fragile point correspondences.
  Simulation across free-form meshes achieved sub-\qty{0.04}{\mm} and sub-\qty{0.4}{\degree} accuracy and robustness to pose noise and contact loss.
  On a tooth-preparation robot, our method attained \qty{0.42}{\mm} and \qty{3.75}{\degree}, outperforming an optical tracker registration while requiring no external sensors.
  These results demonstrate a practical and precise registration strategy for surgical and industrial robots.
\end{abstract}


\section{Introduction}

Accurate registration between a prior model and the real scene is a cornerstone of robotic manipulation.
It anchors task geometry to the robot frame and directly bounds the achievable precision of execution.
Applications such as tooth preparation, orthopedic surgery, and precision assembly demand sub-millimeter accuracy, where even small registration errors can compromise task success.

A widely adopted solution is optical registration, which integrates hand--eye calibration, fiducial detection, and \ac{CAD}-marker manufacturing.
However, long kinematic chains and the accumulation of fabrication and extrinsic errors often limit accuracy even with high-end trackers.
Moreover, line-of-sight constraints restrict usability in confined or occluded workspaces.
These limitations motivate alternatives that bypass external tracking hardware.

Contact-based registration is one such alternative \cite{gunnarsson87cad}.
A rigid probe establishes contact with the object surface, and the robot records the swept trajectory of end-effector poses.
This trajectory is then aligned to the prior model to estimate the object pose.
Yet most existing methods either treat the probe as a point or approximate it with a spherical tip and radius compensation \cite{xiong04nearoptimal,saund17touch,kim23contactbased,haugaard24fixture}.
These simplifications discard geometric constraints imposed by the true probe shape, rely on fragile point correspondences (e.g., \ac{ICP} variants), and degrade when trajectories contain sparse or intermittent contact.
\textbf{This leaves a gap:} a formulation that \emph{explicitly} incorporates probe geometry, leverages both contact and non-contact evidence, and couples efficient global search with metric-grade refinement remains missing.

\begin{figure}
  \centering
  \includegraphics{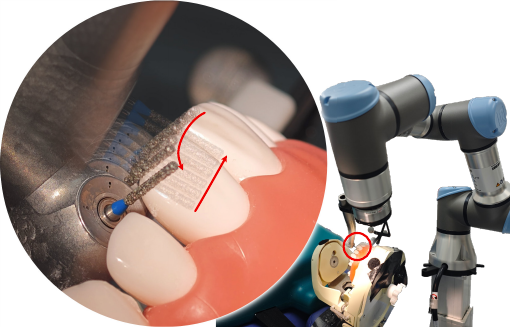}
  \caption{Illustration of proposed registration method in a tooth preparation robot. The manipulator sweeps a cylindrical probe across the target tooth to capture its geometry, which is then registered to a preoperative 3D model for precise motion control.}
  \label{fig:trailer}
\end{figure}

We address this gap by recasting contact registration as a \emph{complementary-shape docking} problem.
The admissible contact set is determined by the probe's full swept volume rather than isolated points (\autoref{fig:trailer}), turning contact from sparse correspondences into dense geometric constraints.
This perspective leverages all trajectory evidence, both contact and non-contact, for robust pose estimation.

Our contributions are:
\begin{itemize}
  \item A reformulation of contact-based registration as complementary-shape docking that encodes probe geometry via swept volumes, avoiding fragile point correspondences.
  \item A two-stage global-to-local search over \(\SE(3)\) that couples FFT-based translation correlation with low-discrepancy orientation sampling, to produce robust initial hypotheses.
  \item A continuous refinement method on \(\SE(3)\) using Lie-algebraic updates and analytic contact sensitivities for precise and stable convergence.
\end{itemize}

The method requires no additional sensing hardware and relies only on manipulator proprioception, making it practical in both medical and industrial settings.
We validated our approach both in simulation and on a tooth-preparation robot using a cylindrical bur as the probe.
In simulation, it achieved an average translational error of \qty{0.031}{\mm} and a rotational error of \qty{0.325}{\degree}.
On the real robot, as shown in \autoref{fig:trailer}, it achieved \qty{0.42}{\mm} translational error and \qty{3.75}{\degree} rotational error, consistently outperforming optical pipelines while requiring no external sensors.

\section{Related Works}

\subsection{Registration Strategies}

Prior work has explored a range of registration strategies, from correspondence-based alignment to global inference.

Early \ac{CMM} pipelines register parts by probing canonical \ac{CAD} features and solving least-squares with assumed correspondences \cite{gunnarsson87cad,li98geometric}.
Grimson and Lozano-P\'erez~\cite{grimson84modelbased} established an early correspondence-driven paradigm by combining sparse points and surface normals with geometric consistency tests and pruning.
Olsson \textit{et al.}~\cite{olsson06registration} showed that correspondence-based registration can be made globally optimal via branch-and-bound over points, lines, and planes, yet it still fundamentally requires reliable correspondences.
Modern contact registration often follows \ac{ICP}-style alignment from sparse touch points and estimated normals, which remains sensitive to outliers and initialization \cite{xiong04nearoptimal,kim23contactbased}.

To reduce dependence on initialization and explicit correspondences, a second line of work reasons globally from contact observations.
Petrovskaya and Khatib \cite{petrovskaya11global} formulate tactile localization as Bayesian inference that treats touch as probabilistic measurements, thereby enabling global pose reasoning.
Haugaard \textit{et al.}~\cite{haugaard24fixture} discretize \(\SE(3)\) and test candidate poses against a \ac{CAD} model, offering correspondence-free global search at the cost of resolution--efficiency trade-offs.
Zhong \textit{et al.}~\cite{zhong23chsel} output diverse pose hypotheses consistent with observed contact and free-space, emphasizing coverage rather than a single metrology-grade estimate.

In summary, correspondence-driven pipelines require accurate matches or well-chosen features and are fragile under sparse or noisy contact, whereas correspondence-free global inference leverages contact and free-space but typically trades precision for tractability.

\subsection{Probe Geometry and Compensation}

How the probe geometry is modeled or compensated has been a central factor in contact-based registration.

A common industrial practice is \ac{CAD} normal offsetting, where the measured contact point is shifted inward by the stylus radius before feature fitting \cite{gunnarsson87cad,li98geometric}.
Ristic \textit{et al.}~\cite{ristic01contact} refine this idea with \ac{CAD} normal compensation coupled with NURBS surface fitting, yet the spherical assumption remains and directional effects are ignored.
Xiong \textit{et al.}~\cite{xiong04nearoptimal} represent the standard metrology view by adopting spherical tips with radius correction in correspondence-driven alignment, which is adequate for ball tips but does not generalize to anisotropic probes.

A second compensation strategy replaces explicit tip handling with a Minkowski sum between the \ac{CAD} surface and a sphere of the probe radius, effectively inflating the object and reducing registration to point--surface alignment on the dilated model \cite{haugaard24fixture}.
Saund \textit{et al.}~\cite{saund17touch} implement an equivalent surrogate by subtracting the probe radius in the SDF test, which approximates a spherical Minkowski sum within a particle-filter framework.

A third line absorbs the probe radius into the measurement noise rather than modeling the geometry explicitly, which simplifies inference but caps achievable accuracy and obscures the distinction between near-contact and penetration \cite{petrovskaya11global}.
Several recent contact-\ac{ICP} systems also neglect radius or treat a sharp ``pointy tool'' as a point, further limiting precision under realistic tool shapes \cite{kim23contactbased,olsson06registration}.
Works that directly assume true surface samples or rely on deformable tactile skins bypass rigid-probe compensation but lie outside our rigid-probe metrology setting \cite{grimson84modelbased,olsson06registration,zhong23chsel}. 

In summary, most pipelines simplify the probe to a sphere and then rely on offsets, Minkowski sums, or noise absorption.
These heuristics ease implementation but blur geometric distinctions and limit achievable accuracy for non-spherical or anisotropic tools.

\section{Preliminaries}

\begin{figure}
  \centering
  \includegraphics{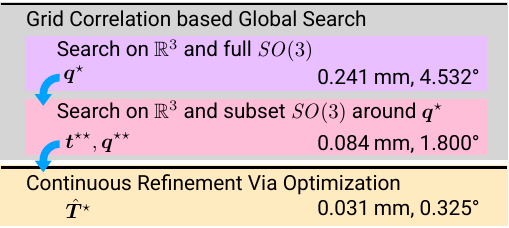}
  \vspace{-1.5em}
  \caption{An overview of the proposed registration pipeline, which consists of three stages: (a) a global search over \(\SE(3)\) with grid correlation and low-discrepancy orientation sampling, (b) a local search on subset \(\SO(3)\) around best orientation \(\vect q^\star\) to improve orientation accuracy, and (c) a continuous optimization on \(\SE(3)\) using Lie-algebraic updates and analytic contact sensitivities for precise convergence. Average error for each stage is shown for reference.}
  \label{fig:pipeline}
\end{figure}

We observe a sequence of probe poses \(\mathcal{P}=\{\matr T_i\}_{i=1}^N\), with \(\matr T_i\in\SE(3)\), describing a trajectory in which a rigid probe attempts to touch the surface of an object. 
Importantly, there is no guarantee that the probe touches the object surface along this trajectory, which makes direct alignment ambiguous.

The task is to estimate a rigid transformation \(\hat{\matr T}\in\SE(3)\) that best aligns the object with the observed trajectory.

To this end, we first perform a global grid-based search with progressively finer orientation sampling.
We then apply a continuous optimization to refine the pose estimate.
The overall procedure is summarized in the three-stage diagram in \autoref{fig:pipeline}.

\section{Grid Correlation based Global Search}

\begin{figure*}
  \centering
  \includegraphics{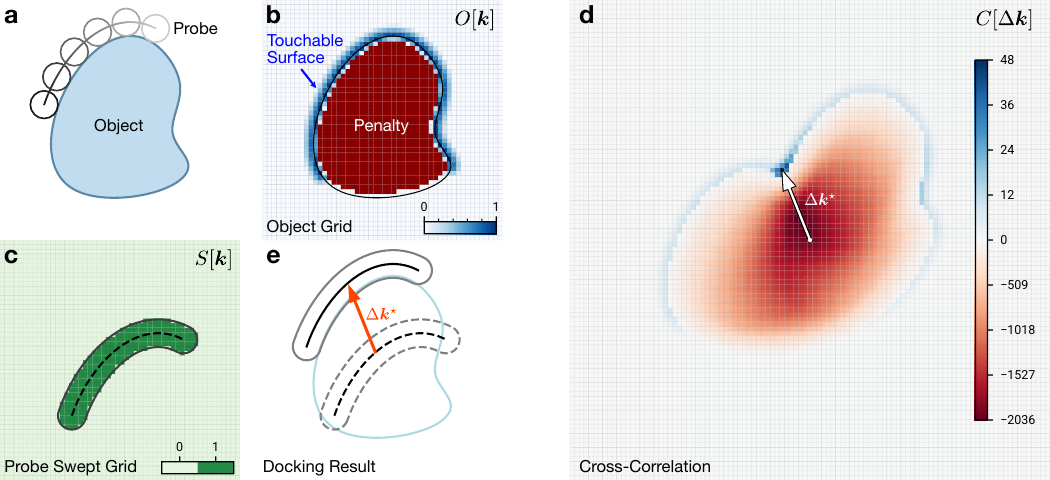}
  \caption{
    Illustration of grid representations and translational search.
    (a) The probe sweeps along the trajectory while in contact, and the object is to be registered.
    (b) Object grid \(O[\vect{k}]\) rewarding near-surface approach, tolerating shallow penetration, and penalizing deep penetration.
    (c) Probe swept grid \(S[\vect{k}]\) indicating all covered voxels along probe's trajectory.
    (d) Cross-correlation \(C[\Delta \vect{k}]\) between the object \(O[\vect{k}]\) and swept volume grids \(S[\vect{k}]\), determining the best translation \(\Delta \vect k^\star\).
    (e) Resulting docking of the probe and object, showing complementary-shape registration respecting contact locality.
  }
  \label{fig:grids}
\end{figure*}

This section describes a two-stage search over \(\SE(3)\): a global exploration followed by a local search within neighborhood.
In the first stage, we evaluate grid-based correlation fields constructed from the object and the probe's swept volume over uniformly sampled orientations to obtain a coarse pose hypothesis.
The resulting coarse orientation is then refined by a second grid-based search constrained to a geodesic ball on \(\SO(3)\), yielding a more accurate candidate.
Our approach is inspired by correlation-based molecular docking \cite{katchalski-katzir92molecular,paggi24art,pagadala17software} , which pioneered the use of grid-based correlation for efficient global search.
In contrast, we adapt the scoring to contact-locality and penetration handling in physical contact and introduce a refinement stage within \(\SE(3)\) to achieve higher accuracy.

\subsection{Object Grid and Probe Swept Volume Grid Representation}

Both the object and the swept volume are discretized into voxel arrays at the same spacing \(h\).
Each array is defined on its own lattice, determined by a limited padding around the \ac{AABB} of the corresponding geometry.
A voxel index \(\vect{k}\in\mathcal{K}\subset\Z^3\) maps to world coordinates \(\vect{x}=h\vect{k}+\vect{o} \in \R^3\) with lattice origin \(\vect{o}\).

\subsubsection{Object Grid}

Two meshes of the object serve complementary roles: a watertight mesh provides inside/outside information, and a non-watertight subset constrains feasible contacts to touchable regions.

From the watertight mesh we compute a \ac{SDF} \(f_o(\vect{x})\), which is mapped into a banded template
\begin{equation}
  O_t(\vect x) =
  \begin{cases}
    1-f_o(\vect x)/\tauoutc & \text{if } 0 \leq f_o(\vect x) < \tauoutc, \\
    1+f_o(\vect x)/\tauinc  & \text{if } -\tauinc < f_o(\vect x) < 0,    \\
    -\rho                   & \text{if } f_o(\vect x) \leq -\tauinc,     \\
    0                       & \text{otherwise},
  \end{cases}
\end{equation}
where \(\tauinc, \tauoutc\) are the inner and outer thresholds, and \(\rho\) is a penalty term for deep penetration.
This rule rewards near-surface approach, tolerates shallow penetration, and penalizes deep penetration, as shown in \autoref{fig:grids}(b).

Most objects have limited surface area that can be contacted by the probe.
For instance, an object placed on a table has only the upper surface reachable.
Such area is represented by a subset of the object mesh, which we call the touchable surface.
From the touchable surface we compute an unsigned distance \(d(\vect{x})\), which measures the distance from the touchable surface.
We define a surface mask \(O_m(\vect x)\) as a linearly feathered band around the zero level set as
\begin{equation}
  O_m(\vect x) = \max\biggl(0, 1 - \frac{\lvert d(\vect x)\rvert}{\tauoutc}\biggr),
\end{equation}
which gates scoring to feasible contact regions.

The final object grid is an element-wise combination
\begin{equation}
  O(\vect x)=\min\bigl(O_t(\vect x),O_m(\vect x)\bigr),
\end{equation}
which simultaneously encodes interior penalties and contact locality.
Sampling on the object lattice yields
\begin{equation}
  O[\vect{k}] = O(h\vect{k}+\vect{o}_o),
\end{equation}
shown in \autoref{fig:grids}(b), where \(\vect o_o\) is the lattice origin.

\subsubsection{Probe Swept Occupancy Grid} 

We construct a binary occupancy field directly from the probe \ac{SDF} and the trajectory.
Let \(f_p(\vect{x})\) denote the probe \ac{SDF} in its local frame, and \(\matr T_i\) be sampled probe poses along trajectory \(\mathcal P\).
Define
\begin{equation}
  S(\vect{x}) = \max_{i=1,\ldots,N} H\bigl(f_p(\matr T_i^{-1}\vect{x})\bigr),
\end{equation}
where \(H\colon \R \mapsto \{0,1\}\) is the Heaviside step function.
Thus \(S(\vect{x})=1\) if the swept volume covers \(\vect{x}\) and \(0\) otherwise.
Discretization on the swept-volume lattice gives
\begin{equation}
  S[\vect{k}] = S(h\vect{k}+\vect{o}_s),
\end{equation}
shown in \autoref{fig:grids}(c), where \(\vect o_s\) is the lattice origin.

\subsection{Translation Search on \texorpdfstring{\(\R^3\)}{ℝ³} via Correlation}
\label{sec:translation-search}

We embed \(S[\vect{k}]\) and \(O[\vect{k}]\) on a common zero-padded canvas at spacing \(h\) without resampling.
Let the embedded arrays be denoted by the same symbols for brevity.
The value is zero if query is outside their supports.

Matching the surfaces is achieved by evaluating the cross-correlation between the two arrays:
\begin{equation}
  C[\Delta\vect{k}] = \sum_{\vect{k}\in\mathcal I_o} O[\vect{k}] S[\vect{k}+\Delta\vect{k}],
  \label{eq:xcorr}
\end{equation}
where \(\mathcal I_o\) is the index set of the object lattice.
We take the maximizer \(\Delta\vect{k}^\star\) as the best voxel shift on the canvas, as shown in \autoref{fig:grids}(d).
The corresponding physical translation between the original geometries is
\begin{equation}
  \vect{t} = h\Delta\vect{k}^\star + \vect{o}_s - \vect{o}_o.
\end{equation}

Direct evaluation of \eqref{eq:xcorr} is computationally heavy, \(\mathcal O(N^6)\).
We accelerate it to \(\mathcal O(N^3\log N)\) with 3D \acp{FFT} using
\begin{equation}
  C = \mathcal{F}^{-1}\bigl(\overline{\mathcal{F}(S)}\odot\mathcal{F}(O)\bigr),
\end{equation}
where \(\odot\) is element-wise multiplication and the overline denotes complex conjugation.
To prevent wrap-around, each axis of the canvas is zero-padded to at least \(n_o+n_s-1\), where \(n_o\) and \(n_s\) are the side lengths of the two finite supports along that axis.

\subsection{Orientation Sampling on \texorpdfstring{\(\SO(3)\)}{SO(3)} and its Subspace}

Exploring translation alone requires a fixed probe orientation, but full 6-DoF alignment demands systematic sampling of orientations on \(\SO(3)\).
Rather than relying on dense grids or random sampling, we adopt a two-stage strategy: a global low-discrepancy sampler for coarse exploration and a novel geodesic ball neighborhood sampler for local refinement.

\subsubsection{Global Orientation Sampling}

For coarse global sampling, we adopt the Super-Fibonacci (SF) sequence~\cite{alexa22superfibonacci}, which generates unit quaternions with low discrepancy directly on the 3-sphere \(\Sph^3\).
Compared to Hopf fibration based method~\cite{yershova10generating}, as used in \cite{haugaard24fixture}, the SF method achieves lower discrepancy and more uniform coverage while remaining equally efficient.

\subsubsection{Geodesic Neighborhood Sampling}

After obtaining the best coarse orientation, we refine the search in its local neighborhood.
We consider the unit quaternion space as the 3-sphere \(\Sph^3\), where the natural distance between two quaternions \(\vect q_1,\vect q_2\) is the half relative rotation angle
\begin{equation}
  d(\vect q_1,\vect q_2) = \arccos\bigl(\lvert \langle \vect q_1,\vect q_2\rangle\rvert\bigr).
\end{equation}
A geodesic ball of radius \(\theta\) centered at quaternion \(\vect q_0\) is thus defined as
\(
  \mathcal{B}(\vect q_0,\theta) = \{ \vect q \in \Sph^3 \colon d(\vect q,\vect q_0)\leq \theta \},
\)
which in rotation space contains all orientations within an angular deviation of \(2\theta\) in the neighbor of \(\vect q_0\).

Uniform sampling in such a restricted region is nontrivial.
The standard SF sequence construction is uniform on the full \(\Sph^3\), but becomes biased when restricted to a subspace.
To preserve uniformity, we replace the radial law with inverse transform sampling with respect to the true radial marginal distribution.
The marginal density of the radius is
\begin{equation}
  \rho(r) \propto rL(r), \quad
  L(r) = 2\arccos\biggl(\frac{\cos\theta}{\sqrt{1-r^2}}\biggr),
\end{equation}
where \(L(r)\) denotes the admissible range of the axial angle at radius \(r\).
The associated cumulative distribution is
\begin{equation}
  F(r) = \frac{1}{Z} \int_0^r \rho(\tilde r)\, \mathrm{d}\tilde r,
\end{equation}
with \(Z\) the normalization constant.
A uniform variable \(u\in[0,1]\) is then mapped to \(r\) by the inverse transform \(r = F^{-1}(u)\).
Since \(F\) has no closed form, we evaluate it numerically on a dense grid, store the tabulated pairs \((r,F(r))\), and recover \(F^{-1}(u)\) via interpolation.
The resulting GeoBall-SF sequence, summarized in \autoref{alg:geoball-sf}, produces low-discrepancy samples uniformly distributed within the geodesic ball.

\vspace{-0.8em}
\begin{algorithm}
  \caption{Generating $n$ samples in a geodesic ball around $\vect q_0$ on $\SO(3)$ (GeoBall-SF)}
  \label{alg:geoball-sf}
  \KwIn{$n,\ \vect q_0,\ \theta,\ \phi,\ \psi,\ \gamma$ \;}
  \KwOut{Unit quaternions $\vect{q}_i$ in $\mathcal{B}(\vect q_0,\theta)$}
  \For{$i \in \{0,\dots,n-1\}$}{
    $s \gets i + 0.5$\;
    $u_1 \gets \frac{s}{\gamma} \bmod 1, u_2 \gets \frac{s}{\phi} \bmod 1, u_3 \gets \frac{s}{\psi} \bmod 1$\;
    $r \gets F^{-1}(u_1),\quad R \gets \sqrt{1-r^2}$\;
    $L \gets 2\arccos(\cos\theta/R)$\;
    $\alpha \gets (u_2-0.5)\,L,\quad \beta \gets 2\pi\,u_3$\;
    $\vect{q}_{\text{local}} \gets (R\cos \alpha,\ R\sin \alpha,\ r\cos\beta,\ r\sin\beta)$\;
    $\vect{q}_i \gets \vect{q}_0 \vect{q}_{\text{local}}$\;
  }
\end{algorithm}

\subsection{Combined Grid-Based Search}

We combine the two components introduced above: \(\R^ 3\) translation search by \ac{FFT} correlation and orientation sampling on \(\SO(3)\).

We sample \(n_g\) quaternions from the global SF sequence. For each quaternion, the object grid \(O\) is rotated into the corresponding orientation and resampled onto the common canvas at spacing \(h\) via trilinear interpolation.
We then compute the correlation \(C\) with the swept-volume occupancy grid \(S\) using \ac{FFT}, recovering the best translation \(\vect{t}\) for that orientation.
Maximizing the score over all orientations and translations yields a coarse pose hypothesis \((\vect t^\star, \vect q^\star)\).

To refine this result, we restrict the search to a geodesic ball on \(\SO(3)\) centered at \(\vect q^\star\).
Candidate quaternions are generated with the GeoBall-SF sampler, ensuring uniform and low-discrepancy coverage within the neighborhood, with sample number \(n_l\) and radius \(\theta\).
For each candidate, the object grid is again rotated, resampled, and correlated with the swept volume, updating the best hypothesis if a higher score is obtained.

In summary, the method performs a global search over \(\SE(3)\) followed by a local refinement in the orientation subspace.
The result is an initial high-quality pose hypothesis \((\vect t^{\star\star}, \vect q^{\star\star})\), which serves as the starting point for the optimization-based refinement in the next section.

\section{Continuous Refinement Via Optimization}

Although the two-stage global search provides a strong hypothesis for the contact registration, it remains limited by the use of discrete orientation sampling and a grid-based field representation, and thus may not fully align the contact geometries.

To address this, we introduce a continuous gradient-based refinement on \(\SE(3)\), optimizing a proximity objective around the best hypothesis \((\vect t^{\star\star}, \vect q^{\star\star})\) through Lie-algebra updates.

\subsection{Objective and Pose Update}

The object pose is represented as \(\matr T=(\matr R, \vect p) \in \SE(3)\) with rotation \(\matr R \in \SO(3)\) and translation \(\vect p \in \R^3\).
We initialize at \(\matr T_\text{init} = (\matr R_\text{init}, \vect t_\text{init})\),
which corresponds to the best hypothesis \((\vect t^{\star\star}, \vect q^{\star\star})\).
The decision variable is a twist
\begin{equation}
  \vect \xi =
  \begin{bmatrix}
    \vect \upsilon &
    \vect \omega
  \end{bmatrix}
  \Tran
  \in \R^6,
\end{equation}
where \(\vect \upsilon\) (mm) and \(\vect \omega\) (rad) are the translational and rotational components.
We map \(\vect \xi\) to an increment \(\increment \matr T=\exp(\vect\xi^\wedge)=(\increment \matr R,\increment \vect p)\) with
\begin{equation}
  \increment \matr R = \exp(\vect \omega^\wedge),
  \quad
  \increment \vect p = \matr V(\vect \omega)\vect \upsilon,
  \label{eq:se3-exp}
\end{equation}
where \(\vect \xi^\wedge \in \se(3)\), \(\vect \omega^\wedge \in \so(3)\) denote the
skew-symmetric matrix representations of the twist \(\vect \xi\) and the rotation vector~\(\vect \omega\).
The \(\matr V(\vect \omega)\) is the \(\SO(3)\) left Jacobian, defined as
\begin{equation}
  \matr V(\vect\omega) = \matr I + \frac{1 - \cos\theta}{\theta^2}\vect\omega^\wedge + \frac{\theta - \sin\theta}{\theta^3}(\vect\omega^\wedge)^2,
\end{equation}
where \(\theta=\lVert\vect\omega\rVert_2\) is the rotation angle.
For \(\theta \to 0\), the series expansion of \(\matr V(\vect\omega)\) is used for numerical stability.

To better decouple the translational and rotational updates, an anchor point is defined in the object body frame as \(\vect a\in\R^3\), which represents the geometry center of the mesh.
We perform a left perturbation about the anchor expressed in the world frame as \(\matr T(\vect\xi) = (\matr R(\vect\xi), \vect p(\vect\xi))\), where
\begin{align}
  \matr R(\vect\xi) & = \increment \matr R\,\matr R_0, \label{eq:left-update-R}                                                  \\
  \vect p(\vect\xi) & = \vect p_0 + \increment \vect p + (\matr I-\increment \matr R)\matr R_0 \vect a. \label{eq:left-update-p}
\end{align}
The compensation term \((\matr I-\increment \matr R)\matr R_0\vect a\) ensures that the rotation is performed about the anchor rather than the world origin.

\begin{figure}
  \centering
  \includegraphics{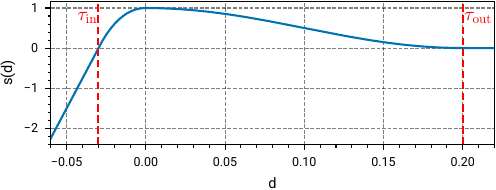}
  \caption{The scoring function \(s(d)\) used in the continuous optimization stage.}
  \label{fig:scoring-plot}
\end{figure}

For each probe pose \(\matr T_i\), we compute the smallest signed distance \(d_i\) between the object and the probe geometry at that pose (see \autoref{sec:min-distance}).
We define a smooth proximity scoring function \(s(d)\) that rewards near-surface proximity while softly penalizing penetration:
\begin{equation}
  t \triangleq \frac{d}{\tauoutr},\quad
  s(d)=
  \begin{cases}
    0,                & t > 1,         \\
    \cos^2(0.5\pi t), & 0 < t \leq 1,  \\
    At^3+Bt^2+1,      & -r < t \leq 0, \\
    k(t+r),           & t \leq -r,
  \end{cases}
\end{equation}
where \(\tauoutr, \tauinr\) denote the outer and inner thresholds,
\(r\triangleq \tauoutr/\tauinr=0.15\) is their ratio, \(k=15\) is the penalty slope,
and the coefficients \(A=k/r^2+2/r^3\), \(B=-k/r-3/r^2\) ensure \(C^1\) continuity.
The resulting function encourages contact points to lie within the template bands while penalizing large deviations (\autoref{fig:scoring-plot}).

The registration maximizes total proximity; equivalently, we minimize
\begin{equation}
  J(\vect\xi)=-\sum_{i=1}^N s\bigl(d_i(\vect\xi)\bigr)+\lambda\bigl(\lVert\vect\upsilon\rVert_2^2+\lVert\vect\omega\rVert_2^2\bigr),
  \label{eq:objective}
\end{equation}
where \(\lambda\geq 0\) is a Tikhonov regularizer.

\medskip
\noindent\textbf{Final Optimization Problem:}
Placing bounds per component yields the constrained problem
\begin{equation}
  \label{eq:reg-problem}
  \begin{aligned}
    \min_{\vect \xi \in \R^6} \quad & J(\vect \xi)                                 \\
    \text{s.t.}\quad                & \lVert\vect\upsilon\rVert_{\infty} \leq b_t, \\
                                    & \lVert\vect\omega\rVert_{\infty} \leq b_r,
  \end{aligned}
\end{equation}

where \(J(\vect\xi)\) is defined in \eqref{eq:objective},
distances \(d_i(\vect\xi)\) are evaluated on the pose \(T(\vect\xi)\) via \eqref{eq:se3-exp}--\eqref{eq:left-update-p},
and \(b_t,b_r\) are tight physical bounds.
We denote the minimizer by \(\vect \xi^\star\);
it defines the refined pose \(\matr T(\vect\xi^\star)\) through \eqref{eq:left-update-R}--\eqref{eq:left-update-p}.

\subsection{Distance Queries on Single Probe Geometry to Mesh}
\label{sec:min-distance}

To evaluate each probe pose in the objective, we require robust queries of the shortest distance between the probe and the target mesh.

The probe shape is represented by a \ac{SDF} \(f_p(\vect x)\).
The mesh is stored in a \ac{BVH}.
For each BVH node, we evaluate \(f_p\) at the node center \(\vect c\) and keep the node if
\begin{equation}
  f_p(\vect c) < r_{\text{node}} + \varepsilon,
\end{equation}
where \(r_{\text{node}}\) is the circumsphere radius of the node and \(\varepsilon\) is a cutoff margin.
This check ensures that triangles with clearance less than \(\varepsilon\) are preserved.
Candidate triangles are further pruned using the same test with their circumsphere. 

For each remaining triangle, a Frank--Wolfe optimization returns a candidate point \(\vect x_k\) and value \(d_k=f_p(\vect x_k)\).
The signed distance is the smallest of these,
\begin{equation}
  d^\star=\min_k d_k ,
\end{equation}
where \(\vect x^\star=\vect x_{k^\star}\) and \(k^\star=\arg\min_k d_k\) is the index of the minimizing triangle.
Negative values indicate penetration, while \(0 \le d^\star < \varepsilon\) indicates positive clearance.
The output is a distance vector
\begin{equation}
  \vect v = -d^\star \vect n^\star,
  \quad
  \vect n^\star = \frac{\nabla f_p(\vect x^\star)}{\lVert \nabla f_p(\vect x^\star) \rVert},
\end{equation}
where \(\vect n^\star\) is the outward \ac{SDF} normal at the closest point.
This vector points from the witness point to the nearest point on the \ac{SDF} surface.

The entire distance query is implemented as a custom \ac{SDF}--mesh collision model within MuJoCo~\cite{todorov12mujoco}.

\subsection{Gradient and Solver}

Let \(\vect v_i, \vect n_i\) be the output from the \ac{SDF} distance query of the \(i\)-th probe pose.
\(\vect v_i = \vect p^i_{\text{obj}}-\vect p^i_{\text{probe}}\), \(\vect n_i\) is the oriented unit normal consistent with the sign of \(d_i\).
Differentials of the signed distance under rigid motion follow standard contact sensitivities:
\begin{equation}
  \frac{\partial d_i}{\partial(\increment \vect p)} = \vect n_i,
  \label{eq:dd-dp}
\end{equation}
\begin{equation}
  \frac{\partial d_i}{\partial \vect\omega}\Big|_{\text{rot}} = (\vect p^i_{\text{obj}}-\vect c)\times \vect n_i,
  \label{eq:dd-domg-rot}
\end{equation}
with pivot \(\vect c = \vect p_0 + \increment \vect p + \matr R_0 \vect a\) from \eqref{eq:left-update-p}.

Since \(\increment \vect p = \matr V(\vect\omega)\vect\upsilon\) by \eqref{eq:se3-exp}, the chain rule to algebra coordinates yields
\begin{align}
  \frac{\partial d_i}{\partial \vect \upsilon}
   & = \matr V(\vect\omega)\Tran \vect n_i,
  \label{eq:dd-du}                                                \\
  \frac{\partial d_i}{\partial \vect \omega}
   & = \bigl(\vect p^i_{\text{obj}}-\vect c\bigr)\times \vect n_i
  +\biggl(\frac{\partial (\matr V(\vect\omega)\vect\upsilon)}{\partial \vect\omega}\biggr)\Tran\vect n_i.
  \label{eq:dd-domg}
\end{align}
The second term in \eqref{eq:dd-domg} captures translation--rotation coupling and is approximated numerically by finite differences.

Let \(\mathrm{d}s/\mathrm{d}d\) denote the derivative of the proximity window.
The full gradient of \eqref{eq:objective} is
\begin{equation}
  \nabla J(\vect\xi) = - \sum_{i=1}^n \frac{\mathrm{d}s}{\mathrm{d}d}\Big|_{d_i(\vect\xi)}
  \begin{bmatrix}
    \partial d_i/\partial \vect\upsilon \\
    \partial d_i/\partial \vect\omega
  \end{bmatrix}
  + 2\lambda
  \begin{bmatrix}
    \vect\upsilon \\
    \vect\omega
  \end{bmatrix}
  .
  \label{eq:gradJ}
\end{equation}

We solve \eqref{eq:objective} with L-BFGS-B optimizer \cite{byrd95limited} on \(\vect \xi\), using initial value \(\vect\xi_0 = \vect 0\), bounds \(\lVert\vect\upsilon\rVert_{\infty} \leq b_t\), \(\lVert\vect\omega\rVert_{\infty} \leq b_r\).
The analytic gradient \eqref{eq:gradJ} is supplied to the optimizer.
We terminate when the projected gradient norm and the step size fall below prescribed thresholds.

\section{Results}

We evaluate the proposed method in both simulation and real-world experiments.
The parameter settings used in all experiments are summarized in \autoref{tab:parameters}.
The average nearest neighbor distance of the global and local orientation samples are \qty{8.0}{\degree} and \qty{0.7}{\degree}, respectively.

\subsection{Simulation}

\begin{table}[t]
  \centering
  \caption{Parameter settings used in all experiments.}
  \label{tab:parameters}
  \begin{tabular}{llll}
    \toprule
    \textbf{Parameter} & \textbf{Value}   & \textbf{Parameter} & \textbf{Value}  \\
    \midrule
    $\rho$             & \num{-50}        & $h$                & \qty{0.2}{\mm}  \\
    $n_g$              & 20000            & $\tauoutc$         & \qty{0.5}{\mm}  \\
    $n_l$              & 5000             & $\tauinc$          & \qty{0.2}{\mm}  \\
    $\theta$           & \qty{5}{\degree} & $\tauoutr$         & \qty{0.2}{\mm}  \\
    $\lambda$          & \num{10e-4}      & $\tauinr$          & \qty{0.03}{\mm} \\
    \bottomrule
  \end{tabular}
  \vspace{-2em}
\end{table}

\begin{figure*}
  \centering
  \includegraphics[width=\linewidth]{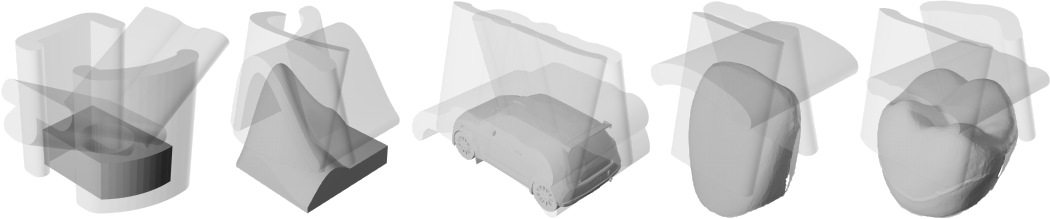}
  \caption{Simulation overview. Contact trajectories were generated with a cylindrical probe on five object meshes. Translucent swept volumes show the simulated probe motion used for registration. Gaussian noise was added during experiments but omitted here for clarity.}
  \label{fig:sim-overview}
\end{figure*}

\begin{table*}
  \centering
  \caption{Simulation results}\label{tab:sim-results}
  \begin{tabular}{cccccccc}
    \toprule
    \multirow{2}*{Mesh Name} & \multirow{2}*{\makecell{Trajectory \\ Length}} & \multicolumn{2}{c}{Global Search Results} & \multicolumn{2}{c}{Local Search Results} & \multicolumn{2}{c}{Refinement Stage Results} \\
    \cmidrule(lr){3-4}  \cmidrule(lr){5-6}  \cmidrule(lr){7-8}
                &                          & Trans. Err(\unit{\mm}) & Rot. Err(\unit{\degree}) & Trans. Err(\unit{\mm}) & Rot. Err(\unit{\degree}) & Trans. Err(\unit{\mm}) & Rot. Err(\unit{\degree}) \\
    \midrule
    Workpiece   & 395                      & 0.089\,\textpm\,0.007  & 3.117\,\textpm\,0.000    & 0.102\,\textpm\,0.007  & 2.750\,\textpm\,0.000    & 0.021\,\textpm\,0.003  & 0.214\,\textpm\,0.042    \\
    Matlab Logo & 525                      & 0.368\,\textpm\,0.011  & 3.117\,\textpm\,0.000    & 0.073\,\textpm\,0.039  & 2.163\,\textpm\,0.513    & 0.040\,\textpm\,0.004  & 0.512\,\textpm\,0.112    \\
    Toy Car     & 542                      & 0.177\,\textpm\,0.033  & 3.117\,\textpm\,0.000    & 0.068\,\textpm\,0.006  & 1.678\,\textpm\,0.979    & 0.034\,\textpm\,0.003  & 0.322\,\textpm\,0.099    \\
    Front Teeth & 242                      & 0.107\,\textpm\,0.075  & 5.290\,\textpm\,0.000    & 0.101\,\textpm\,0.031  & 1.110\,\textpm\,0.254    & 0.030\,\textpm\,0.003  & 0.403\,\textpm\,0.235    \\
    Molar Teeth & 339                      & 0.463\,\textpm\,0.009  & 8.018\,\textpm\,0.000    & 0.078\,\textpm\,0.012  & 1.299\,\textpm\,0.511    & 0.032\,\textpm\,0.001  & 0.174\,\textpm\,0.041    \\
    \midrule
    \multicolumn{2}{c}{Mean Across Meshes} & 0.241\,\textpm\,0.166  & 4.532\,\textpm\,2.164    & 0.084\,\textpm\,0.016  & 1.800\,\textpm\,0.666    & 0.031\,\textpm\,0.007  & 0.325\,\textpm\,0.138    \\
    \bottomrule
  \end{tabular}
\end{table*}

We generated contact trajectories in simulation to evaluate the proposed method, using a cylindrical probe with a diameter of \qty{1.4}{\mm} and a length of \qty{20}{\mm}.

Five object meshes were used in the simulation, as shown in \autoref{fig:sim-overview}:
a workpiece (simple geometric shape suitable for traditional \ac{CMM} registration),
the Matlab logo (mathematical free-form surface),
a toy car (\ac{CAD}-derived object with complex geometry),
and front and molar teeth (meshes scanned by an intraoral scanner).
All meshes were scaled so that their longest side measured approximately \qtyrange{5}{8}{\mm}.

For each object, we generated a contact trajectory via OpenCAMLib \cite{wallinopencamlib}, simulating the probe contacting the object surface.
Each trajectory consisted of three segments with different contact directions.

To better mimic real-world execution, we perturbed each trajectory by adding Gaussian noise with a standard deviation of \qty{0.02}{\mm} independently to each pose sample.
The trajectory was sampled to be as realistic as possible for a robot to execute using a force--position hybrid controller.
The trajectories are visualized in \autoref{fig:sim-overview}.

The calibration results were evaluated by the translational error of the mesh geometry center and the rotation error.
\autoref{tab:sim-results} reports the results of three stages.
The global search stage achieved an average translation error of \qty{0.241}{\mm} and rotation error of \qty{4.532}{\degree}.
The local search stage achieved an average translation error of \qty{0.084}{\mm} and rotation error of \qty{1.800}{\degree}.
The refinement stage further improved the accuracy to \qty{0.031}{\mm} and \qty{0.325}{\degree}, demonstrating a significant benefit.
The average runtime was \qty{115}{\second}.

\begin{figure}
  \centering
  \includegraphics{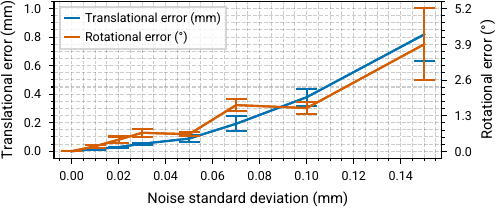}
  \vspace{-0.8em}
  \caption{Noise robustness evaluation. The plot shows the registration error under different levels of Gaussian noise added to the poses.}
  \label{fig:sim-noise}
\end{figure}

To evaluate robustness, we added independent Gaussian noise to each pose in the trajectory.
The noise variance was varied across experiments, and the results are shown in \autoref{fig:sim-noise}.
The proposed method remained robust up to \qty{0.1}{\mm} noise: below this threshold, both translational and rotational errors remained small and the simulation results were stable.
When the noise standard deviation exceeded \qty{0.1}{\mm}, accuracy dropped sharply and the simulation results became unstable.
This failure occurs when a large portion of the swept volume intrude into the mesh, causing heavy penetration penalties and negative scores during local refinement.

\begin{figure}
  \centering
  \includegraphics{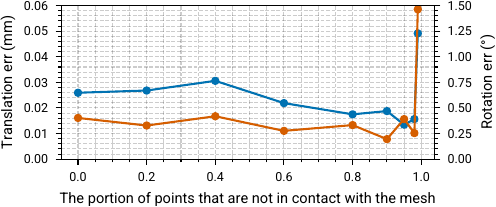}
  \vspace{-0.8em}
  \caption{Sparse contact evaluation. The registration remained accurate with up to \qty{98}{\percent} of the trajectory not touching the surface.}
  \label{fig:sim-parttouch}
\end{figure}

In addition to the Gaussian noise experiments, we also evaluated the method under sparse contact conditions, where only a portion of the trajectory points were in actual contact with the mesh.
This behavior reflects realistic scenarios, where the object may ``hop'' or intermittently touch the surface rather than maintaining continuous contact.
Remarkably, the registration remained accurate even with up to \qty{98}{\percent} of the trajectory not touching the surface, as shown in \autoref{fig:sim-parttouch}.
The results indicate that the method can achieve reliable registration with sparse data, demonstrating robustness in practical scenarios.

\subsection{Real-World Experiments}

Since acquiring ground truth for real-world registration is difficult, the accuracy is indirectly assessed through the outcomes of tooth preparation.

We evaluated the proposed method on a tooth preparation robot system, consisting of a UR3e manipulator equipped with a dental handpiece with a cylindrical bur.
Its task is to perform tooth preparation surgery on a phantom teeth, shaping the target tooth according to a predesigned model.
Such surgery requires an accurate model registration between the preoperative 3D model and the real tooth.

The accurate tooth mesh was obtained by a high-accuracy intraoral scanner (3Shape TRIOS).
The cylindrical bur acted as a probe, with its diameter of \qty{1}{\mm} and length precisely defined.
The tool center point (TCP) was calibrated using the four-point method provided by the UR controller.

We used a force-position hybrid controller to maintain a contact force of \qty{0.3}{\newton} during sweep contact.
The contact path was planned similar to the simulated trajectories, with three segments of different contact directions.
The trajectory execution time was \qty{80}{\second}.
After the registration, the tooth preparation was executed using the obtained results.

\begin{figure}
  \centering
  \includegraphics{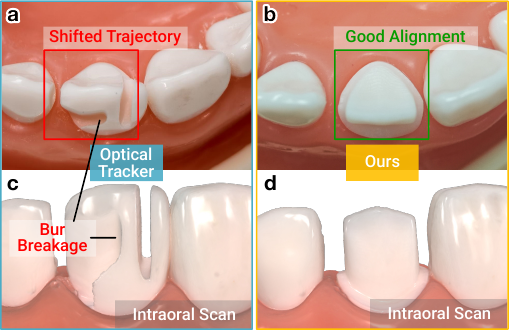}
  \vspace{-2em}
  \caption{Comparison of the prepared tooth using the optical tracker and the proposed registration method.
    (a) With the optical tracker, the trajectory was shifted leftward, damaging adjacent teeth, and the bur broke mid-procedure when forced to remove excess material caused by the misalignment. (b) Using our method, the prepared tooth closely matched the designed shape, demonstrating substantially improved registration accuracy.}
  \label{fig:ndivscc}
  \vspace{-1em}
\end{figure}

For comparison, we also evaluated an optical-tracker-based calibration chain.
Specifically, we used an NDI Polaris Lyra optical tracking system, which is widely adopted for high-precision pose measurement, with a reported error of \qty{0.40}{\mm} at the \qty{95}{\percent} confidence level and a volumetric RMS accuracy of \qty{0.20}{\mm} \cite{northerndigitalinc23polaris}.
In our implementation, the NDI setup used a standard hand--eye calibration (\(AX=XB\)) to determine the robot--NDI transformation.
The tooth--NDI relationship is obtained using a custom guide plate with attached markers.
The guide is \ac{CAD}-designed based on the intraoral scan and fabricated by \qty{25}{\um}-accuracy 3D printing, but it still introduced noticeable printing and mounting errors.
Errors in this chain led to trajectory shifts: during experiments, the robot damaged both the neighboring teeth and the target tooth, and excessive cutting even broke the bur (see \autoref{fig:ndivscc}).

In contrast, our method requires no such calibration chain, and the preparation closely matched the planned tooth shape without damaging adjacent teeth or overloading the bur.

To evaluate its accuracy, we used an indirect method: the unprepared teeth served as stable references, since their geometry remains unchanged during preparation.
Using these neighbors, we aligned the designed tooth model (from the initial intraoral scan) with the postoperative oral scan using \ac{ICP}, treating this as an approximate ground truth between the two oral scans.
We also performed \ac{ICP} using the prepared tooth itself as the reference, assuming the prepared shape ideally matched the design except for a rigid offset.

\begin{figure}
  \centering
  \includegraphics{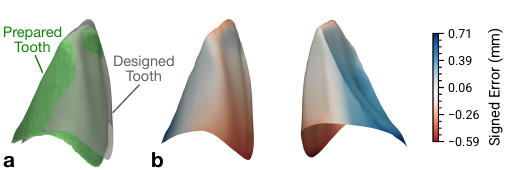}
  \caption{Comparison of designed tooth and prepared tooth using the proposed registration method. (a) Mesh visualization. (b) Per-point distance error between designed mesh and prepared tooth mesh.} 
  \label{fig:mesh_error}
  \vspace{-1em}
\end{figure}

By comparing the transformation from the approximate ground truth with the transformation estimated by our registration method, we obtained a pose error of \qty{3.75}{\degree} in rotation and \qty{0.42}{\mm} in translation.
Furthermore, we computed the per-point distance error between the designed mesh and the prepared tooth mesh (\autoref{fig:mesh_error}).
The distance error ranged from \qty{-0.59}{\mm} (over-cut) to \qty{0.71}{\mm} (under-cut).

These results confirm that our registration method produces more reliable and accurate preparations than the optical-tracker-based approach, achieving better adherence to the planned tooth geometry.

\section{Conclusions}

We proposed a calibration-free, probe-aware registration method that reformulates alignment as complementary-shape docking between the object and the probe's swept volume.
The method achieved average \qty{0.03}{\mm} and \qty{0.32}{\degree} accuracy in simulation and \qty{0.42}{\mm} and \qty{3.75}{\degree} in real world experiments, outperforming an optical calibration chain while requiring only manipulator proprioception.
In practice, the method delivers high precision on free-form surfaces while requiring neither external sensors nor careful initialization, making it a practical registration solution for surgical and industrial robots operating in constrained or occluded workspaces.
A limitation is the reliance on an accurate object model, and future work will explore confidence evaluation and the use of contact force cues to further strengthen robustness.

\FloatBarrier
\bibliographystyle{IEEEtran}
\bibliography{dental}

\end{document}